\documentclass[runningheads]{llncs}
\usepackage{graphicx}
\usepackage{comment}
\usepackage{amsmath,amssymb} %
\usepackage{color}

\usepackage[width=122mm,left=12mm,paperwidth=146mm,height=193mm,top=12mm,paperheight=217mm]{geometry}
\usepackage{multirow}
\usepackage{booktabs}
\usepackage[pagebackref=true,breaklinks=true,letterpaper=true,colorlinks,linkcolor=blue,citecolor=blue,bookmarks=false]{hyperref}
\usepackage{xspace}
\usepackage{bbm}
\usepackage{array}
\usepackage{breakcites}
\usepackage{colortbl}
\usepackage{lipsum}

\definecolor{Gray}{gray}{0.9}

\makeatletter
\DeclareRobustCommand\onedot{\futurelet\@let@token\@onedot}
\def\@onedot{\ifx\@let@token.\else.\null\fi\xspace}

\def\ie{\emph{i.e}\onedot, } 
 
 \def\vs{\emph{vs}\onedot}
 
\def\etal{\emph{et al}\onedot}
\def\vs{\emph{vs}\onedot }
\makeatother

\newcommand{\cmmnt}[1]{}

\newcommand{\netname}{RSC-Net}

\begin{document}
\pagestyle{headings}
\mainmatter
\def\ECCVSubNumber{787}  %

\title{3D Human Shape and Pose from \\ a Single Low-Resolution Image  \\ with Self-Supervised Learning} %
%
%

%

%
%

%
%
\titlerunning{Low-resolution 3D human shape and pose}
\author{Xiangyu Xu\inst{1} \and
Hao Chen\inst{2} \and
Francesc Moreno-Noguer\inst{3} \index{Moreno-Noguer, Francesc} \and \\
L{\'a}szl{\'o} A. Jeni\inst{1} \index{Jeni, L{\'a}szl{\'o} A.} \and
Fernando De la Torre\inst{1,4} \index{De la Torre, Fernando}
}
\authorrunning{X. Xu et al.}
\institute{
Robotics Institute, Carnegie Mellon University, Pittsburgh, USA \and
Electrical and Computer Engineering, Carnegie Mellon University, Pittsburgh, USA \and
Institut de Rob\`{o}tica i Inform\`{a}tica Industrial (CSIC-UPC), Barcelona, Spain \and
Facebook Reality Labs (Oculus), Pittsburgh, USA
}
\maketitle

\begin{abstract}
	3D human shape and pose estimation from monocular images has been an active area of research in computer vision, having a substantial impact on the development of new applications, from activity recognition to creating virtual avatars. 
	Existing deep learning methods for 3D human shape and pose estimation rely on relatively high-resolution input images; however, high-resolution visual content is not always available in several practical scenarios such as video surveillance and sports broadcasting.
	Low-resolution images in real scenarios can vary in a wide range of sizes, and a model trained in one resolution does not typically degrade gracefully across resolutions. 
	Two common approaches to solve the problem of low-resolution input are applying super-resolution techniques to the input images which may result in visual artifacts, or simply
	training one model for each resolution, which is impractical in many realistic applications. 

	To address the above issues, this paper proposes a novel algorithm called \netname{}, which consists of a {R}esolution-aware network, a {S}elf-supervision loss, and a {C}ontrastive learning scheme. 
	The proposed network is able to learn the 3D body shape and pose across different resolutions with a single model.
	The self-supervision loss encourages scale-consistency of the output, and the contrastive learning scheme enforces scale-consistency of the deep features. 
	We show that both these new training losses provide robustness when learning 3D shape and pose in a weakly-supervised manner. 
	Extensive experiments demonstrate that the \netname{} can achieve consistently better results than the state-of-the-art methods for  challenging low-resolution images. %
\end{abstract}

\section{Introduction}
\begin{figure}[t]
	\centering
\includegraphics[width=0.85\textwidth]{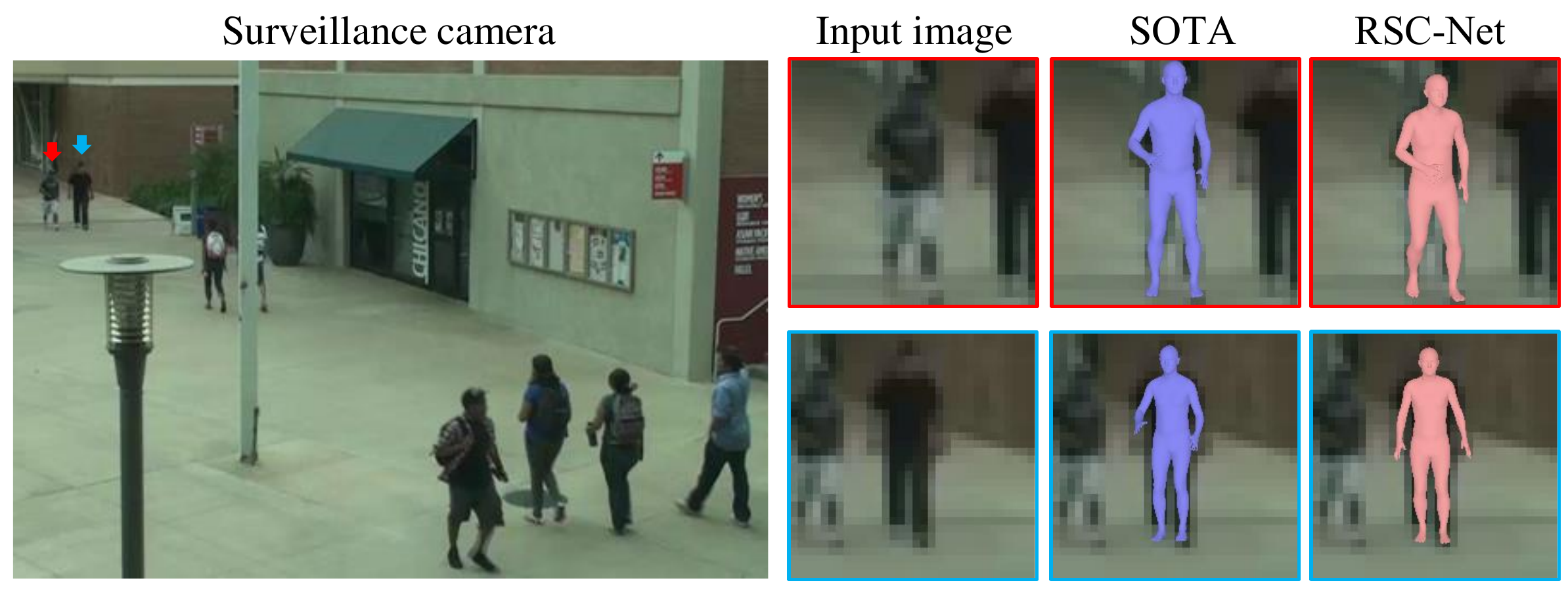} 
	\caption{%
		3D human shape and pose estimation from a low-resolution image captured from a real surveillance video. SOTA
	method~\cite{kolotouros2019spin} that works well for high-resolution images performs poorly at low-resolution ones.
	}
	\label{fig:teaser_real}
\end{figure}

3D human shape and pose estimation from 2D images is of great interest to the computer vision and graphics community.
Whereas significant progress has been made in this field,
it is often assumed that the input image is high-resolution and contains sufficient information for reconstructing the 3D human geometry in detail~\cite{alldieck2019learning,alldieck2018video,bogo2016keep,kanazawa2018end,kanazawa2019learning,kocabas2019vibe,kolotouros2019spin,natsume2019siclope,pavlakos2018learning,pumarola20193dpeople,saito2019pifu,zheng2019deephuman}.
However, this assumption does not always hold in practice, since lots of images in real scenes have low resolutions, such as surveillance cameras and sports videos~\cite{xxy-iccv17,wang2016studying,nishibori2014exemplar,neumann2018tiny,xu2019towards,oh2011large}.
As a result, existing algorithms designed for high-resolution images are prone to fail when applied to low-resolution inputs as shown in   Figure~\ref{fig:teaser_real}. 
In this paper, we study the relatively unexplored problem  of estimating  3D human shape and pose from  low-resolution images.
There are two major challenges of this low-resolution 3D estimation problem.
First, the resolutions of the input images in real scenarios vary in a wide range, and a network trained for one specific resolution does not always work well for another.
One might consider overcoming this problem by simply training different models, one for each image resolution. 
However, this is impractical in terms of memory and training computation.
Alternatively, one could super-resolve the images to a sufficiently large resolution, but the super-resolution step often results in visual artifacts, which leads to poor 3D estimation.
To address this issue, we propose a resolution-aware deep neural network for 3D human shape and pose estimation that is robust to different image resolutions. 
Our network builds upon two main components: a feature extractor shared across different resolutions and a set of resolution-dependent parameters to adaptively integrate the different-level features.

Another challenge we encounter is due to the fact that high-quality 3D annotations are hard to obtain, especially for in-the-wild data, and only a small portion of the training images have 3D ground truth labels~\cite{kanazawa2018end,kolotouros2019spin}, which complicates the training process.
Whereas most training images have 2D keypoint labels, they are usually not sufficient for predicting the 3D outputs due to the inherent ambiguities in the 2D-to-3D mapping.
This problem is further accentuated in our task, as the low-resolution 3D estimation is not well constrained and has a large solution space due to limited pixel observations.
Therefore, directly training low-resolution models with incomplete information typically does not achieve good results.
Inspired by the self-supervised learning~\cite{laine2017temporal,tarvainen2017mean}, we propose a directional self-supervision loss to remedy the above issue.
Specifically, we enforce the consistency across the outputs of the same input image with different resolutions, such that the results of the higher-resolution images can act as guidance for lower-resolution input. This strategy significantly improves the 3D estimation results.

In addition to enforcing output consistency, we also devise an approach to enforce consistency of the feature representations across different resolutions. 
Nevertheless, we find that the commonly used mean squared error is not effective in measuring discrepancies between high-dimensional feature vectors. 
Instead, we adapt the contrastive learning~\cite{oord2018representation,he2019momentum,chen2020simple} which aims to maximize the mutual information across the feature representations at different resolutions, and encourages the network to produce better features for the low-resolution input.

To summarize, we make the following contributions in this work.
First, we study the relatively unexplored problem of 3D human shape and pose estimation from low-resolution images and present a simple yet effective solution for it, called \netname{}, which is based on a novel resolution-aware network that can handle arbitrary-resolution input with one single model. 
Second, we propose a self-supervision loss to address the issue of weak supervision.
Furthermore, we introduce contrastive learning  which effectively enforces the feature consistency across different resolutions.
Extensive experiments demonstrate that the proposed method outperforms the state-of-the-art algorithms on challenging low-resolution inputs and achieves robust performance for high-quality 3D human shape and pose estimation.
\section{Related Work}
We first review the state-of-the-art methods for 3D human shape and pose estimation and then discuss the low-resolution image recognition algorithms.

%
%
%
\noindent \textbf{3D human shape and pose estimation.}
Recent years have witnessed significant progress in the field of 3D human shape and pose estimation from a single image~\cite{alldieck2019learning,alldieck2018video,alldieck2019tex2shape,bogo2016keep,doersch2019sim2real,kanazawa2018end,kanazawa2019learning,kocabas2019vibe,kolotouros2019spin,natsume2019siclope,pavlakos2018learning,pumarola20193dpeople,saito2019pifu,zheng2019deephuman,zhang2019predicting,zanfir2018monocular}.
Existing methods for this task can be broadly categorized into two classes.
The first kind of approaches generally splits the 3D human estimation process into two stages: first transforming the input image into new representations, such as human 2D keypoints~\cite{bogo2016keep,pavlakos2018learning,natsume2019siclope,alldieck2018video,alldieck2019learning,doersch2019sim2real}, human silhouettes~\cite{pavlakos2018learning,alldieck2018video,natsume2019siclope}, body part segmentations~\cite{alldieck2019learning}, UV mappings~\cite{alldieck2019tex2shape}, and optical flow~\cite{doersch2019sim2real}, and then regressing the 3D human parameters~\cite{loper2015smpl} from the transformed outputs of the last stage either with iterative optimization~\cite{bogo2016keep,alldieck2018video} or neural networks~\cite{pavlakos2018learning,alldieck2019learning,doersch2019sim2real,natsume2019siclope}.
As these methods map the original input images into simpler representation forms which are generally sparse and can be easily rendered, they can exploit a large amount of synthetic data for training where there are sufficient high-quality 3D labels.
However, these two-stage systems are error-prone, as the errors from early stage may be accumulated or even deteriorated~\cite{kanazawa2018end}. 
In addition, the intermediate results may throw away valuable information in the image such as context.
More importantly, the task of the first stage, \ie to estimate the intermediate representations, is usually difficult for low-resolution images, and thereby, the aforementioned two-stage models are not suitable to solve our problem of low-resolution 3D human shape and pose estimation.

Without relying on new representations, the second kind of approaches can directly regress the 3D parameters from the input image~\cite{kanazawa2018end,kanazawa2019learning,kolotouros2019spin,saito2019pifu,kocabas2019vibe,pumarola20193dpeople,zhang2019predicting}, where most methods are based on deep neural networks.
While being concise and not requiring the estimation of intermediate results, these methods usually suffer from the problem of weak supervision due to a lack of high-quality 3D ground truth.
Most existing works focus on this problem and have developed different techniques to solve it. 
As a typical example, Kanazawa \etal \cite{kanazawa2018end} include a generative adversarial network (GAN)~\cite{gan} to constrain the solution space using the prior learned from 3D human data.  
However, we find the GAN-based algorithm less effective for low-resolution input images where substantially fewer pixels are available.
Kolotouros~\cite{kolotouros2019spin} \etal integrate the optimization-based method \cite{bogo2016keep} into the training process of the deep network to more effectively exploit the 2D keypoints. 
While achieving good improvements over~\cite{kanazawa2018end} on high-resolution images, \cite{kolotouros2019spin} cannot be easily applied to low-resolution input, as the low-resolution network cannot provide good initial results to start the optimization loop. 
In addition, it significantly increases the training time.
On the other hand, temporal information has also been exploited to enforce temporal consistency of the 3D estimation results, which however requires high-resolution video input~\cite{kanazawa2019learning,zhang2019predicting,kocabas2019vibe}.
Different from the above methods, we propose a 3D human shape and pose estimation algorithm using a single low-resolution image as input.
We propose self-supervision loss and contrastive feature loss which effectively remedy the problem of insufficient 3D supervision.

%
%
%
%
%
%
%
%
%
%
%
%
%
%
%
%
%
%
%
%
%
%
%
%

%
%
%
\noindent \textbf{Low-resolution image recognition.}
While there is no prior work for low-resolution 3D human shape and pose estimation, there are some related approaches to process low-resolution inputs for other image recognition tasks, such as 2D body pose estimation~\cite{neumann2018tiny}, face recognition~\cite{ge2018low,cheng2018low,xxy-iccv17}, image classification~\cite{wang2016studying}, image retrieval ~\cite{tan2018feature,Noh2019better}, and object detection~\cite{haris2018task,li2017perceptualgan}.
Most of these methods address the low-resolution issue by enhancing the degraded input, in either the image space~\cite{haris2018task,cheng2018low,wang2016studying} or the feature space~\cite{ge2018low,tan2018feature,li2017perceptualgan,Noh2019better}.
One typical image-space method~\cite{haris2018task} applies a super-resolution network which is trained to improve both the image quality (\ie per-pixel similarity such as PSNR) and the object detection performance.
However, the loss functions for higher PSNR and better recognition performance do not always agree with each other, which may lead to inferior solutions. Moreover, the super-resolution model may bring unpleasant artifacts, resulting in domain gap between the super-resolved and real high-resolution images.
Unlike the image enhancement based approaches, the feature enhancement based methods~\cite{ge2018low,tan2018feature,li2017perceptualgan,Noh2019better} are not distracted by the image quality loss and thus can better focus on improving the recognition performance.
As a representative example, Ge \etal~\cite{ge2018low} use mean squared error (MSE) to enforce the similarity between the features of low-resolution and high-resolution images, which achieves good results for face recognition.
Different from the above approaches, Neumann~\etal~\cite{neumann2018tiny} propose a novel method for low-resolution 2D body pose estimation by predicting a probability map with Gaussian Mixture Model,
which, however, cannot be easily extended to 3D human shape and pose estimation.
In this work, we apply the feature enhancement strategy to low-resolution 3D human shape and pose estimation. 
Instead of using MSE for measuring feature similarity, we introduce the contrastive learning~\cite{oord2018representation} which can more effectively maximize the mutual information across the features of different resolutions.
In addition, we handle different-resolution input with a resolution-aware neural network.

\begin{figure}[t]
	\begin{center}
		\begin{tabular}{c}
			\includegraphics[width = 0.9\linewidth]{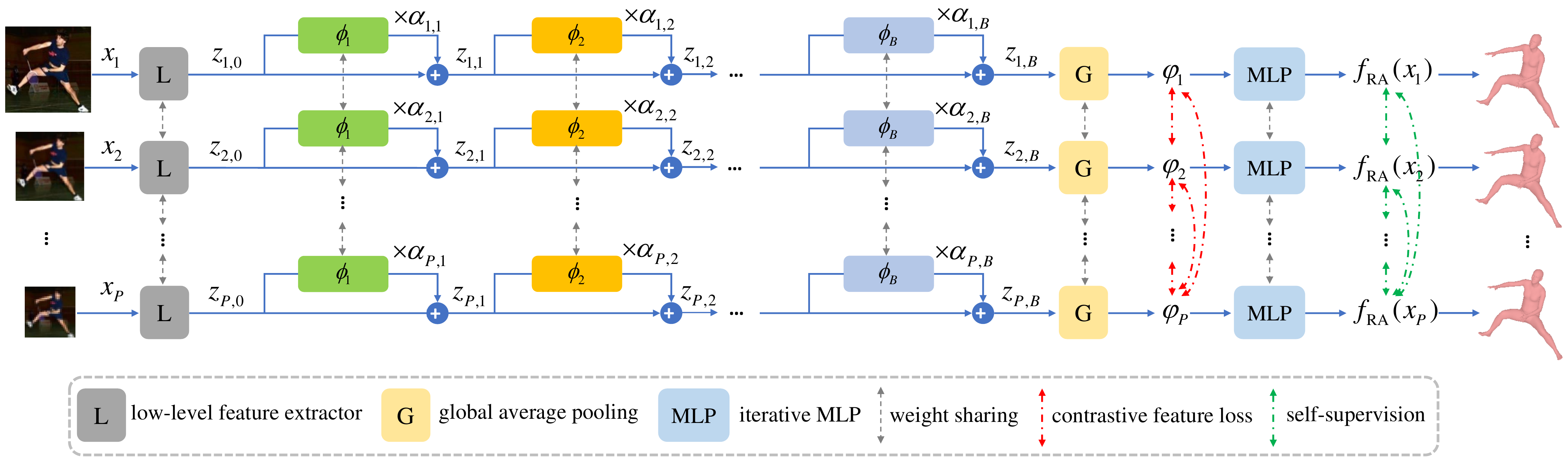} \\
		\end{tabular}
	\end{center}
	\caption{Overview of the proposed algorithm. The resolution-aware network $f_{\text{RA}}$ is trained with a combination of the basic loss (omitted in the figure for simplicity), self-supervision loss and contrastive feature loss. The modules with the same colors are shared across different resolutions, while the matrix $\alpha$ is resolution-dependent. Note that we resize  the different resolution inputs $\{x_i\}$ to $224\times224$ with bicubic interpolation before feeding them into the network.
	}
	\label{fig:resolution-aware network}
\end{figure}
\section{Algorithm}
We study the problem of 3D human shape and pose estimation for a low-resolution image $x$.
Instead of training different networks for each specific resolution, we propose a resolution-aware neural network $f_{\text{RA}}$ which can handle the complex inputs with different resolutions.
We first introduce the 3D human representation model and the baseline network for 3D human estimation with a single 2D image.
Then we describe the proposed resolution-aware model as well as the self-supervision loss and the contrastive learning strategy for training the network.
An overview of our method is shown in Figure~\ref{fig:resolution-aware network}.

\subsection{3D Human Representation}
We represent the 3D human body using the Skinned Multi-Person Linear (SMPL) model~\cite{loper2015smpl}.
The SMPL is a parametric model which describes the body shape and pose with two sets of parameters $\beta$ and $\theta$, respectively. 
The body shape is represented by a basis in a low-dimensional shape space learned from a training set of 3D human scans, and the parameters $\beta \in \mathbb{R}^{10}$ are coefficients of the basis vectors.
The body pose is defined by a skeleton rig with $K=24$ joints including the body root, and the pose parameters $\theta\in \mathbb{R}^{3K}$ are the axis-angle representations of the relative rotation between different body parts as well as the global rotation of the body root.
%
%
%
With $\beta$ and $\theta$, we can obtain the 3D body mesh: $M = f_{\text{SMPL}}(\beta, \theta)$, where $M \in \mathbb{R}^{N \times 3}$ is a triangulated surface with $N=6890$ vertices.

Similar to the prior works~\cite{kanazawa2018end,kolotouros2019spin}, we can predict the 3D locations of the body joints $X$ with the body mesh using a pretrained mapping matrix $W \in \mathbb{R}^{K \times N}$:
\begin{align} \label{eq:3D_joints}
X \in \mathbb{R}^{K \times 3} = WM.
\end{align}
With the 3D human joints, we use a perspective camera model to project the body joints from 3D to 2D. Assuming the camera parameters are $\delta \in \mathbb{R}^{3}$ which define the 3D translation of the camera, the 2D keypoints can be formulated as:
\begin{align} \label{eq:2D_joints}
J \in \mathbb{R}^{K \times 2} = f_{\text{project}}(X, \delta),
\end{align}
where $f_{\text{project}}$ is the perspective projection function~\cite{hartley2003multiple}. %

\subsection{Resolution-Aware 3D Human Estimation} \label{sec:res-aware}
\noindent \textbf{Baseline network.}
Similar to the existing methods~\cite{kanazawa2018end,kolotouros2019spin}, we use the deep convolutional neural network (CNN) for 3D human estimation, where the ResNet-50~\cite{resnet} is employed to extract features from the input image. 
The building block of the ResNet (\ie ResBlock \cite{he2016identity}) can be formulated as:
\begin{align} \label{eq:resblock}
z_{k} &= z_{k-1} + \phi_{k}(z_{k-1}), 
\end{align}
where $z_{k}$ is the output features of the $k$-th ResBlock, and $\phi_k$ represents the nonlinear function used to learn the feature residuals, which is modeled by several convolutional layers with ReLU activation~\cite{nair2010rectified}.
The ResNet stacks $B$ ResBlocks together, and the final output can be written  as: 
\begin{align}\label{eq:resnet}
z_{B} &= z_{0} + \sum_{k=1}^B \phi_{k}(z_{k-1}),
\end{align}  
where $z_0$ is the low-level features extracted from the input image $x$ with convolutional layers, and $z_B$ is a combination of different level residual maps from all the ResBlocks. 
Note that we do not explicitly consider the downsampling ResBlocks in \eqref{eq:resnet} for clarity.
With the output features of the ResNet, we can use global average pooling to obtain a feature vector $\varphi$ and employ an iterative MLP for regressing the 3D parameters $\beta, \theta, \delta$ similar to \cite{kanazawa2018end,kolotouros2019spin}.

%
%
%
%
%
%

%
%

%
\noindent \textbf{Resolution-aware network.}
The baseline network is originally designed for high-resolution images with input size 224$\times$224 pixels,
whereas the image resolutions for human in real scenarios can be much lower and vary in a wide range. 
A straightforward way to deal with these low-resolution inputs is to train different networks for all possible resolutions and choose the suitable one for each test image.
However, this is impractical for real applications. 

To solve this problem, we propose a resolution-aware network, and the main idea is that the different-resolution images with the same contents are largely similar as shown in Figure~\ref{fig:resolution-aware network} and can share most parts of the feature extractor. And only a small amount of parameters are needed to be resolution-dependent to account for the characteristics of different image resolutions. 
Towards this end, instead of directly combining the different level features as in \eqref{eq:resnet},  
we learn a matrix $\alpha$ to adaptively fuse the residual maps from the ResBlocks for each input resolution as shown in Figure~\ref{fig:resolution-aware network}, such that different resolutions can have suitable features for 3D estimation.
Specifically, we formulate the output of the proposed resolution-aware network as:
\begin{align} \label{eq:res_aware_net}
z_{i,B} &= z_{i,0} + \sum_{k=1}^B \alpha_{i,k} \phi_{k}(z_{i,k-1}),~~~i=1,2,\dots,R,
\end{align}
where $i$ is the index for different image resolutions, and larger $i$ indicates smaller image. $i=1$ corresponds to the original high-resolution input. $\alpha \in \mathbb{R}^{R \times B}$, where $R$ denotes the number of all the image resolutions considered in this work. 
$z_{i,k}$ and $\alpha_{i,k}$ respectively represent the output and the fusion weight of the $k$-th ResBlock for the $i$-th input resolution.
According to \eqref{eq:res_aware_net}, the original ResBlock in \eqref{eq:resblock} is modified as: $z_{i,k} = z_{i,k-1} + \alpha_{i,k}\phi_{k}(z_{i,k-1})$.
Note that we use a slightly different notation here compared with \eqref{eq:resblock} and \eqref{eq:resnet} which do not have the index $i$ for image resolution, as the baseline network is not resolution-aware and applies the same operations to different resolution inputs.

Note that for training the above network, each high-resolution image in the training dataset needs to conduct the downsampling operation for $R-1$ times, such that each row of parameters in $\alpha$ have their corresponding training data.
Whereas the original training datasets~\cite{3dpw,human3.6,mpi-inf-3dhp,mpii,coco} are already quite large for the diversity of the training images, it will be further augmented by $R-1$ times, which significantly increases the computational burden of the training process.
To remedy the training issues and reduce the parameters in $\alpha$, we divide all the $R$ resolutions %
into $P$ ranges and only learn one set of parameters for each range. 
We design the first resolution range to only have the original high-resolution image, and for the other ranges, we randomly sample a resolution in each range during each training iteration.
The training images with different resolutions can be denoted as $\{x_i, i=1,2,\dots, P\}$ where the smaller images $x_2, x_3, \dots, x_P$ are synthesized from the same high-resolution image $x_1$ with bicubic interpolation. 
With this strategy, the training set can be much smaller without losing diversity, and we can have a lower-dimensional matrix $\alpha \in \mathbb{R}^{P \times B}$, where the number of parameters can be reduced from $RB$ to $PB$. 
During inference, we first decide the resolution range of the input image and then choose the suitable row of parameters in $\alpha$ for usage in the network. 

%
%
\noindent \textbf{Progressive training.}
Directly using different resolution images for training all at once can lead to difficulties in optimizing the proposed model since the network needs to handle inputs with complex resolution properties simultaneously.
Instead, we train the proposed network in a progressive manner, where the higher-resolution images are easier to handle and thus first processed in training, and more challenging ones with lower resolutions are subsequently added.
In this way, we alleviate the difficulty of the training process and the proposed model can evolve progressively.
%
%

%
%
\noindent \textbf{Basic loss function.}
Similar to the previous algorithms~\cite{kanazawa2018end,kolotouros2019spin}, the basic loss of our network is a combination of 3D and 2D losses.
Suppose the output of the proposed network for input image $x_i$ is $ [ \hat{\beta}_i, \hat{\theta}_i, \hat{\delta}_i ] = f_{\text{RA}}(x_i)$ where $i$ is the resolution index, and ${X_g, J_g, \beta_{\text{g}}}, {\theta_{\text{g}}}$ are the ground truth 3D and 2D keypoints and SMPL parameters.
The basic loss function can be written as:
\begin{align}\label{eq:basic loss}
L_{\text{b}} =\sum_i \| [\hat{\beta}_i, \hat{\theta}_i] - [{\beta_{\text{g}}}, {\theta_{\text{g}}}] \|_2^2 + \lambda_1 \| \hat{X}_i - X_{\text{g}} \|_2^2 +\lambda_2 \| \hat{J}_i - J_{\text{g}} \|_2^2, 
\end{align}
where $\hat{X}_i$ and $\hat{J}_i$ are estimated with \eqref{eq:3D_joints} and \eqref{eq:2D_joints}, respectively. $\lambda_1$ and $\lambda_2$ are hyper-parameters for balancing different terms.
Note that while all the training images have 2D keypoint labels $J_g$ in  \eqref{eq:basic loss}, only a limited portion of them have 3D ground truth $X_g, \beta_g, \theta_g$.
For the training images without 3D labels, we simply omit the first two terms in \eqref{eq:basic loss} similar to \cite{kanazawa2018end,kanazawa2019learning,kolotouros2019spin}

\subsection{Self-Supervision}

The 3D human shape and pose estimation is a weakly-supervised problem as only a small part of the training data has 3D labels, and it is especially the case for in-the-wild images where accurate 3D annotations cannot be easily captured.
This issue gets even worse for the low-resolution images, as the 3D estimation is not well constrained by limited pixel observations, which requires strong supervision signal during training to find a good solution.

To remedy this problem, we propose a self-supervision loss to assist the basic loss for training the resolution-aware network $f_\text{RA}$. 
This new loss term is inspired by the self-supervised learning algorithm~\cite{laine2017temporal} which improves the training by minimizing the MSE between the network predictions under different input augmentation conditions.
For our problem, we naturally have the same input with different data augmentations, \ie the different-resolution images synthesized from the same high-resolution image. 
Thus, the self-supervision loss can be formulated by enforcing the consistency across the outputs of different image resolutions:
\begin{align} \label{eq:self-supervision-v1}
\sum_{i, j}	\|f_{\text{RA}}(x_i)-f_{\text{RA}}(x_j)\|_2^2.
\end{align}

However, a major difference between our work and the original self-supervision method~\cite{laine2017temporal} is that we are generally more confident in the predictions of the higher-resolution images while \cite{laine2017temporal} treats the results under different input augmentations equally.
To exploit this prior knowledge, we improve the loss in \eqref{eq:self-supervision-v1} and propose a directional self-supervision loss:
\begin{align}
\begin{aligned} \label{eq:self-supervision-v2}
 L_{\text{s}} =& \sum_{i,j} w_{i,j}	\|\bar{f}_{\text{RA}}(x_i)-f_{\text{RA}}(x_j)\|_2^2, \\
 w_{i,j} &= \mathbbm{1}(j-i>0) \cdot (j-i),
\end{aligned}
\end{align}
where $w_{i,j}$ is the loss weight for an image pair $(x_i, x_j)$, and it is nonzero only when $x_i$ has higher-resolution than $x_j$.
$\bar{f}_{\text{RA}}$ represents a fixed network, and the gradients are not back-propagated through it such that the lower-resolution image $x_j$ is encouraged to have similar output to higher-resolution $x_i$ but not vice versa.
In addition, since higher-resolution results usually provide higher-quality guidance during training, we give a larger weight to larger resolution difference by the term $(j-i)$ in $w_{i,j}$.
Note that we use all the resolutions that are higher than $x_j$ as supervision in \eqref{eq:self-supervision-v2} instead of only using the highest resolution $x_1$, as the results of $x_j$ and $x_1$ can differ from each other significantly for a large $j$, and the results of the resolutions between $x_j$ and $x_1$ can act as soft targets during training. In \cite{hinton2015distilling}, Hinton \etal show the effectiveness of the ``dark knowledge'' in soft targets, and similarly for low-resolution 3D human shape and pose estimation, we also find that it is important to provide the challenging input a hierarchical supervision signal such that the learning targets are not too difficult for the network to follow. 

\subsection{Contrastive Learning}
While the self-supervision loss enforces the consistency of the network outputs across different image resolutions, we can further improve the model training by encouraging the consistency of the final feature representation $\varphi$ encoded by the network, such that features of lower-resolution images are closer to those of higher-resolution ones.
Similar to \eqref{eq:self-supervision-v2}, we have the feature consistency loss:
\begin{align}
	L_{\text{f}} = \sum_{i,j} w_{i,j} g(\bar{\varphi}_i, \varphi_j),
\end{align}
where $\varphi_{i}$ is the feature vector of the $i$-th resolution input image $x_i$, and $\bar{\varphi}$ denotes a fixed feature extractor without gradient back-propagation. 
$w_{i,j}$ is identical to that in \eqref{eq:self-supervision-v2}.
The function $g$ is used to measure the distance between two feature vectors, and a straightforward choice is the MSE as in \eqref{eq:self-supervision-v2}.
However, the extracted features $\varphi$ usually have very high dimensions, and the MSE loss is not effective in modeling correlations of the complex structures in high-dimensional representations, due to the fact that it can be decomposed element-wisely, \ie assuming independence between elements in the feature vectors~\cite{oord2018representation,tian2019contrastive}.
Moreover, the unimodal losses such as MSE can be easily affected by the noise or insignificant structures in the features, while a better loss function should exploit more global structures~\cite{oord2018representation}.

Towards this end, we propose a contrastive feature loss similar to \cite{oord2018representation,chen2020simple,he2019momentum,tian2019contrastive} to maximize the mutual information across the feature representations of different resolutions.
The main idea behind our contrastive loss is to encourage the feature representation to be close for the same image with different resolutions but far for different images. 
Mathematically, the contrastive function can be written as:
\begin{align}\label{eq:contrastive}
g(\bar{\varphi_{i}}, \varphi_{j}) = -\log \frac{\exp(s(\bar{\varphi_{i}}, \varphi_{j}) / \tau)}{\exp(s(\bar{\varphi_{i}}, \varphi_{j}) / \tau) + \sum_{q \in \mathcal{Q}} \exp(s(q, \varphi_{j}) / \tau) },
\end{align}
where $s$ represents the cosine similarity function, and $\tau$ is a temperature hyper-parameter.
$\varphi_{i}, \varphi_{j}$ are the features of the same input with different resolutions. 
$\mathcal{Q}$ is a queue of data samples, which is constructed and progressively updated during training, and $\varphi_i, \varphi_j \notin \mathcal{Q}$.
We use a method similar to \cite{he2019momentum} to update the queue, \ie after each iteration, the current mini-batch is enqueued, and the oldest mini-batch in the queue is removed.
Supposing the size of the queue is $|\mathcal{Q}|$,
the contrastive loss is essentially a $(|\mathcal{Q}|+1)$-way softmax-based classifier which classifies different resolutions $(\varphi_i, \varphi_j)$ as a positive pair while different contents $(q, \varphi_j)$ as a negative pair.
As the feature extractor of the higher resolution image does not have gradients in \eqref{eq:contrastive}, the proposed loss function enforces the network to generate higher-quality features for the low-resolution input image. 

Our final loss is a combination of the basic loss, self-supervision loss, and  contrastive feature loss: $L_{\text{b}} + \lambda_\text{s} L_\text{s} + \lambda_\text{f} L_\text{f}$, where $\lambda_\text{s}$ and $\lambda_\text{f}$ are hyper-parameters.
\section{Experiments}
We first describe the implementation details of the proposed \netname{}.
Then we compare our results with the state-of-the-art 3D human estimation approaches for different image resolutions.
We also perform a comprehensive ablation study to demonstrate the effect of our contributions.
\subsection{Implementation Details}
We train our model and the baselines using a combination of 2D and 3D datasets similar to previous works \cite{kanazawa2018end,kolotouros2019spin}. 
For the 3D datasets, we use Human3.6M \cite{human3.6} and MPI-INF-3DHP \cite{mpi-inf-3dhp} with ground truth of 3D keypoints, 2D keypoints, and SMPL parameters.
These datasets are mostly captured in constrained environments, and models trained on them do not generalize well to diverse images in real world.
For better performance on in-the-wild data, we also use the 2D datasets including LSP~\cite{lsp}, LSP-Extended \cite{lsp-extend}, MPII \cite{mpii}, and MS COCO \cite{coco}, which only have 2D keypoint labels. 
We crop the human regions from the images and resize them to 224$\times$224.
Images with significant occlusions or small human are discarded from the dataset.
We consider human image resolutions ranging from 224 to 24.
As introduced in Section~\ref{sec:res-aware}, we split all the resolutions into $P=5$ ranges: $\{224, (224, 128], (128, 64], (64, 40], (40, 24]\}$, where the first range corresponds to the original high-resolution image $x_1$.
We obtain the lower-resolution images by downsampling the high-resolution images and resize them back to 224 with bicubic interpolation.
During training, we apply data augmentations to the images including Gaussian noise, color jitters, rotation, and random flipping.
For the loss functions, we set $\lambda_1=5$, $\lambda_2=5$, $\lambda_\text{s}=0.1$, and $\lambda_\text{f}=0.1$.
For contrastive learning, we set the size of the queue as $8192$ and $\tau=0.1$ in \eqref{eq:contrastive} similar to \cite{chen2020simple}.
As in \cite{kocabas2019vibe}, we initialize the baseline networks and our model with the parameters of \cite{kolotouros2019spin}.
We use the Adam algorithm~\cite{kingma2014adam} to optimize the network with a learning rate 5e-5.
Similar to \cite{kocabas2019vibe}, we conduct evaluations on a large in-the-wild dataset 3DPW~\cite{3dpw} with 3D joint ground truth to demonstrate the strength of our model in an in-the-wild setting.
We also provide results for constrained indoor images using the MPI-INF-3DHP dataset~\cite{mpi-inf-3dhp}. 
Following \cite{kocabas2019vibe,kanazawa2018end,kolotouros2019spin}, we compute the procrustes aligned mean per joint position error (MPJPE-PA) and mean per joint position error (MPJPE) for measuring the 3D keypoint accuracy. 
To evaluate the performance of different image resolutions, we report results for the middle point of each resolution range, \ie 176, 96, 52, and 32. 

\begin{figure}[t]
	\centering
	\includegraphics[width=0.92\textwidth]{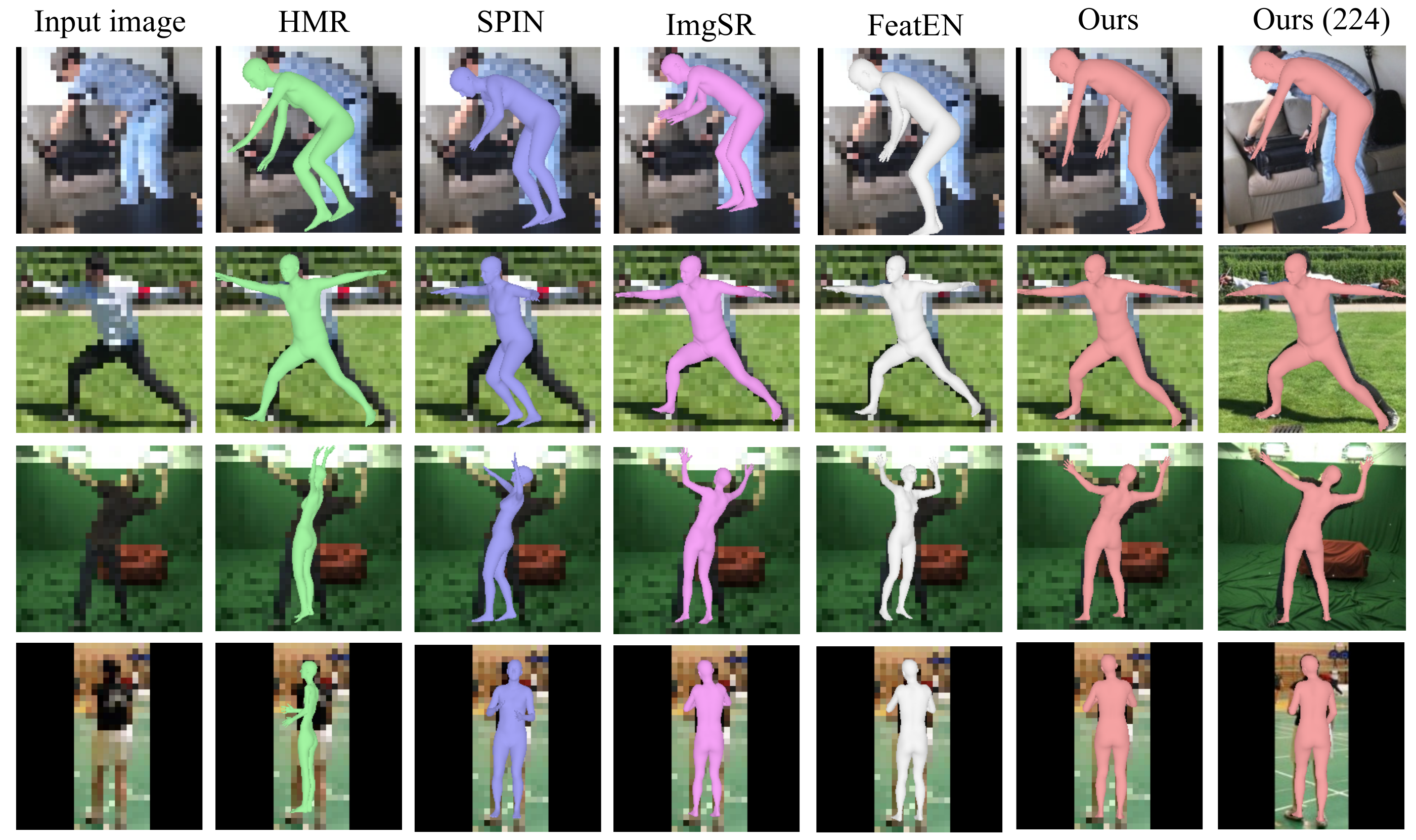} 
	\caption{Visual comparisons with the state-of-the-art methods on challenging low-resolution input. The input image has a resolution of $32\times32$. The results of high-resolution images are also included as a reference. All the baselines are trained with the same training data as our method. }
	\label{fig:SOTA comparison}
\end{figure}

\begin{table}[t]
	\centering
	\caption{\label{tab:compare_3dpw} Quantitative evaluations against the state-of-the-arts on 3DPW~\cite{3dpw}.}
	\begin{tabular}{l*{4}{>{\centering\arraybackslash}p{0.095\textwidth}} >{\centering\arraybackslash}p{0.005\textwidth} *{4}{>{\centering\arraybackslash}p{0.095\textwidth}}}
		\toprule
		\multirow{2}{*}{Methods~~} & \multicolumn{4}{c}{MPJPE} & & \multicolumn{4}{c}{MPJPE-PA} \\
		\cmidrule{2-5}  \cmidrule{7-10} 
		& 176  & 96  & 52  & 32 &  &  176  & 96  & 52  & 32  \\
		\midrule
		HMR        &   117.86  &  118.91 & 125.95  & 142.29 &   &   70.28   &    70.89  &   73.64    & 79.73     \\
		SPIN        &   112.72 &  113.60  & 120.71  & 137.61  &    &  69.20  & 69.40 & 72.21  & 78.44    \\	
		ImgSR  &     116.47  & 117.74 &  127.78  & 146.58  &    &  66.62    &  67.48 & 72.34 & 81.07   \\
		FeaEN  &     107.97  &   109.42  & 119.08  & 143.51 &     &   61.37      &  62.13   &   66.62  & 77.21 \\
		Ours  &  \bf 96.36   &  \bf 97.36    & \bf 103.49   &  \bf 117.12  &   & \bf 58.98 & \bf 59.34   & \bf 61.81 & \bf 67.59 \\
		\bottomrule
	\end{tabular}
\end{table}

\begin{table}[t]
	\centering
	\caption{\label{tab:compare_mpi-inf} Quantitative evaluations against the state-of-the-arts on MPI-INF-3DHP~\cite{mpi-inf-3dhp}.}
	\begin{tabular}{l*{4}{>{\centering\arraybackslash}p{0.095\textwidth}} >{\centering\arraybackslash}p{0.005\textwidth} *{4}{>{\centering\arraybackslash}p{0.095\textwidth}}}
		\toprule
		\multirow{2}{*}{Methods~~} & \multicolumn{4}{c}{MPJPE} & & \multicolumn{4}{c}{MPJPE-PA} \\
		\cmidrule{2-5}  \cmidrule{7-10} 
		&  176  & 96  & 52  & 32 &  & 176  & 96  & 52  & 32  \\
		\midrule
		HMR        &    114.89  &  113.27 & 114.82  & 133.25 &   &     74.77    &   74.45  &   76.35    & 85.30     \\
		SPIN        &  108.46 &  108.25  &  113.36   & 127.27  &    &   71.19  & 71.53 & 74.76 &   83.38   \\		
		ImgSR  &     107.98  & 107.56 &  112.14  & 125.91   &    &   72.13     &  72.76 & 75.64 & 83.52     \\
		FeaEN &     110.40   & 109.91    & 113.09  & 124.99 &  &  71.49  & 71.52    & 73.92    & 81.80    \\
		Ours  &    \bf 103.36   &\bf 103.39     & \bf 106.04  &  \bf 115.80  &   &   \bf 70.01 &   \bf 70.27   & \bf 72.56     & \bf 78.68     \\
		\bottomrule
	\end{tabular}
\end{table}

\subsection{Comparison to State-of-the-Art Methods}
We compare against the state-of-the-art 3D human shape and pose estimation methods HMR~\cite{kanazawa2018end} and SPIN~\cite{kolotouros2019spin} by fine-tuning them on different resolution images with the same training settings as our model.
Since no previous approach has focused on the problem of low-resolution 3D human shape and pose estimation, we adapt the low-resolution image recognition algorithms to our task as new baselines, including both image super-resolution based~\cite{haris2018task} and feature enhancement based~\cite{tan2018feature}.
For the image super-resolution based method (denoted as ImgSR), we first use a state-of-the-art network RDN~\cite{SR-residual-dense} to super-resolve the low-resolution image, and the output is then fed into SPIN~\cite{kolotouros2019spin} for regressing the SMPL parameters. 
Similar to \cite{haris2018task}, the network is trained to improve both the perceptual image quality and the 3D human shape and pose estimation accuracy.
For feature enhancement (denoted as FeaEN), we apply the strategy in \cite{tan2018feature} which uses a GAN loss to enhance the discriminative ability of the low-resolution features for better image retrieval performance.
Nevertheless, we find the WGAN~\cite{arjovsky2017wasserstein} used in the original work \cite{tan2018feature} does not work well in our experiments, and we instead use the LSGAN~\cite{mao2017least} combined with the basic loss~\eqref{eq:basic loss} to train a stronger baseline network.

As shown in Table~\ref{tab:compare_3dpw} and \ref{tab:compare_mpi-inf}, the proposed method compares favorably against the baseline approaches on both 3DPW and MPI-INF-3DHP datasets for all the image resolutions. 
Note that we achieve significant improvement over the baselines on the 3DPW dataset as shown in Table~\ref{tab:compare_3dpw}, which demonstrates the effectiveness of the proposed method on the challenging in-the-wild images.
We also provide a qualitative comparison against the baseline models in Figure \ref{fig:SOTA comparison}, where the proposed method generates higher-quality 3D human estimation results on the challenging low-resolution input. 
\begin{table}[t]
	\centering
	\caption{\label{tab:ablation} Ablation study of the proposed method. Ba: baseline network with basic loss function, RA: resolution-aware network with basic loss function, SS: self-supervision loss, MS: MSE feature loss, CD: cosine distance feature loss, CL: contrastive learning feature loss.}
	%
	\begin{tabular}{l*{4}{>{\centering\arraybackslash}p{0.095\textwidth}} >{\centering\arraybackslash}p{0.005\textwidth} *{4}{>{\centering\arraybackslash}p{0.095\textwidth}}}
		\toprule
		\multirow{2}{*}{Methods} & \multicolumn{4}{c}{MPJPE} && \multicolumn{4}{c}{MPJPE-PA} \\
		\cmidrule{2-5}  \cmidrule{7-10} 
		&  176  & 96  & 52  & 32 & &  176  & 96  & 52  & 32  \\
		\midrule
		Ba                 &    112.26 & 115.18 &  124.88  & 143.63 &  & 65.04  & 66.41 &   71.12  &  79.43   \\
		Ba+SS           &  107.51 & 109.58  & 116.54     &  128.88   &  &   62.32 & 63.27  &  66.78 & 72.49   \\
		RA                &    111.55  &  112.18   &  118.70   & 135.29  & &   64.53   & 68.88  &  68.01 &  75.49    \\
		RA+SS        &   102.56 & 104.18 & 110.17     & 124.23   &  &  60.17  &  60.84  & 63.71  & 69.87  \\
		RA+SS+MS          &  105.96 & 106.15 & 111.33 & 124.85  & &    60.90 &  61.76 &  64.55 & 70.40  \\
		RA+SS+CD        &   104.95 & 105.96  & 111.41 &  125.08 & &  61.29 & 61.91 & 64.30 & 70.17 \\
		RA+SS+CL~         &  \bf 96.36   &  \bf 97.36    & \bf 103.49   &  \bf 117.12  &   & \bf 58.98 & \bf 59.34   & \bf 61.81 & \bf 67.59 \\ 
		\bottomrule
	\end{tabular}
\end{table}

\subsection{Ablation Study}
\begin{figure}[t]
	\centering
	\includegraphics[width=0.9\textwidth]{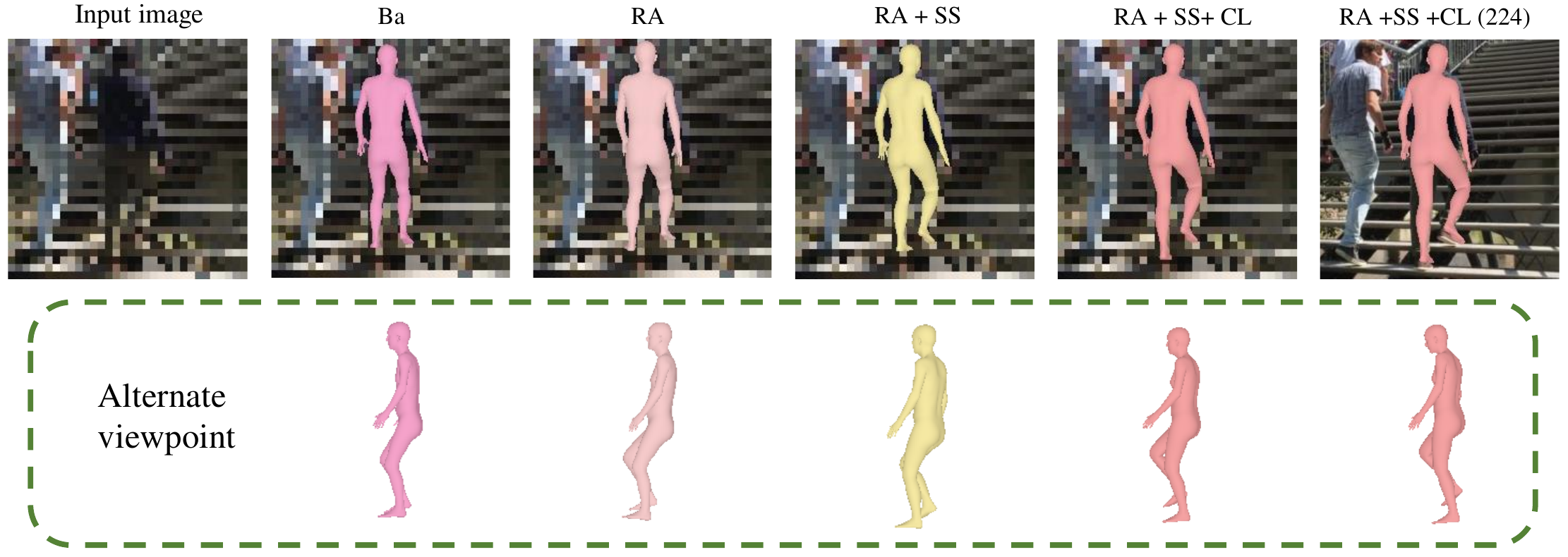} 
	\caption{Visual example which shows the effectiveness of the resolution-aware network, the self-supervision loss, and the contrastive learning feature loss.}
	\label{fig:visual_ablation}
\end{figure}

We provide an ablation study using the 3DPW dataset in Figure~\ref{fig:visual_ablation} and Table~\ref{tab:ablation} to evaluate the proposed resolution-aware network, self-supervision loss, and contrastive feature loss.
We first compare the proposed resolution-aware network with the baseline model ResNet50~\cite{kanazawa2018end,resnet}. 
As shown by ``RA'' and ``Ba'' in Table \ref{tab:ablation}, our network can obtain slightly better results than the baseline network with the basic loss~\eqref{eq:basic loss} as loss function.
Further, we can achieve a more significant improvement over the baseline when adding the self-supervision loss~\eqref{eq:self-supervision-v2} for training, \ie ``RA+SS'' \vs ``Ba+SS'', which further demonstrates the effectiveness of the resolution-aware structure.

Second, we use the self-supervision loss in~\eqref{eq:self-supervision-v2} to exploit the consistency of the outputs of the same input image with different resolutions.
By comparing ``RA+SS'' against ``RA'' in Table \ref{tab:ablation}, we show that the self-supervision loss is important for addressing the weak supervision issue of  3D human pose and shape estimation and thus effectively improves the results. 
The comparison between ``Ba+SS'' and ``Ba'' also leads to similar conclusions.

In addition, we propose to enforce the consistency of the features across different image resolutions.
However, a normally-used MSE loss does not work well as show in ``RA+SS+MS'' of Table \ref{tab:ablation}, which is mainly due to that the unimodal losses are not effective in modeling the correlations between high-dimensional vectors and can be easily affected by noise and insignificant structures in the embedded features~\cite{oord2018representation}. 
In contrast, the proposed contrastive feature loss can more effectively improve the feature representations by maximizing the mutual information across the features of different resolutions, and achieve better results as in ``RA+SS+CL'' of Table \ref{tab:ablation}.
Note that we adopt the cosine similarity in the contrastive feature loss~\eqref{eq:contrastive} similar to prior methods~\cite{oord2018representation,he2019momentum,tian2019contrastive}. 
Alternatively, one may only use the cosine distance function for measuring the distance of two features instead of using the whole contrastive loss~\eqref{eq:contrastive}.
Nevertheless, this strategy does not work well as shown by ``RA+SS+CD'' in Table \ref{tab:ablation}, which demonstrates the effectiveness of the proposed algorithm.

%

%
%
%
%
%
%
%
%
%
%
%
%
%
%
%
%
%
%
%
%
%

%
\noindent \textbf{Analysis of training strategies.}
We also provide a detailed analysis of the alternative training strategies of our model. 
First, as described in Section~\ref{sec:res-aware}, we train our model as well as the baselines in a progressive manner to deal with the challenging multi-resolution input.
As shown in the first row of Table \ref{tab:analysis} (\ie ``w/o PT''), directly training the model for all image resolutions without the progressive strategy leads to degraded results.

Second, the original self-supervision loss~\eqref{eq:self-supervision-v1} treats the images under different augmentations equally, while we are generally more confident in the high-resolution predictions. 
Therefore, we propose a directional self-supervision loss in \eqref{eq:self-supervision-v2} to exploit this prior knowledge.
As shown in the second row of Table \ref{tab:analysis} (\ie ``w/ SS-o''), using the original self-supervision loss~\eqref{eq:self-supervision-v1} is not able to achieve high-quality results, as the network can minimize \eqref{eq:self-supervision-v1} by simply degrading the high-resolution predictions without improving the results of low resolution. 
In addition, we provide hierarchical supervision for low-resolution images in \eqref{eq:self-supervision-v2} which can act as soft targets during training.
As shown in Table \ref{tab:analysis}, only using the highest-resolution predictions as guidance (\ie ``w/ SS-h'') cannot produce as good results as the proposed approach (\ie ``full model'').

\begin{table}[t]
	\centering
	\caption{\label{tab:analysis} Analysis of the alternative training strategies. PT: Progressive Training, SS-o: original self-supervision loss, SS-h: only using the highest-resolution for supervision.}
	\begin{tabular}{l*{4}{>{\centering\arraybackslash}p{0.095\textwidth}} >{\centering\arraybackslash}p{0.005\textwidth} *{4}{>{\centering\arraybackslash}p{0.095\textwidth}}}
		\toprule
		\multirow{2}{*}{Methods~~} & \multicolumn{4}{c}{MPJPE} & & \multicolumn{4}{c}{MPJPE-PA} \\
		\cmidrule{2-5}  \cmidrule{7-10} 
		&  176  & 96  & 52  & 32 &  &  176  & 96  & 52  & 32  \\
		\midrule
		w/o PT                &     105.11     &  106.60   & 113.41  & 127.05 &  &    61.46    &  62.22   &  65.47   & 71.30    \\
		w/ SS-o           & 143.31  & 142.32  &  145.61 &  156.25  &  &  77.75 & 77.51 & 79.06  &  82.97  \\
		w/ SS-h            &  104.16     & 105.24  &  109.94    &  122.01   &  &     62.46  &  62.73   &  64.47   & 68.89 \\
		full model       &     \bf 96.36   &  \bf 97.36    & \bf 103.49   &  \bf 117.12  &   & \bf 58.98 & \bf 59.34   & \bf 61.81 & \bf 67.59 \\
		\bottomrule
	\end{tabular}
\end{table}

\section{Conclusion}
In this work, we  study the challenging problem of low-resolution 3D human shape and pose estimation and present an effective solution, the \netname{}.
We propose a resolution-aware neural network which can deal with different resolution images with a single model.
For training the network, we propose a directional self-supervision loss which can exploit the output consistency across different resolutions to remedy the issue of lacking high-quality 3D labels.
In addition, we introduce a contrastive feature loss which is more effective than MSE for measuring high-dimensional vectors and helps learn better feature representations. 
Our method performs favorably against the state-of-the-art methods on different resolution images and achieves high-quality results for low-resolution 3D human shape and pose estimation.

\bibliographystyle{splncs04}
\bibliography{egbib}
\end{document}